\pdfoutput=1
\documentclass[11pt]{article}

\usepackage[]{emnlp2021}

\usepackage{times}
\usepackage{latexsym}
\usepackage{graphicx}
\usepackage{subfigure}
\usepackage{booktabs}

\usepackage[T1]{fontenc}
\usepackage[utf8]{inputenc}

\usepackage{microtype}

\title{\textsc{Ernie-M}: Enhanced Multilingual Representation by Aligning Cross-lingual Semantics with Monolingual Corpora}

\author{Xuan Ouyang, Shuohuan Wang, Chao Pang, Yu Sun, \\
\textbf{Hao Tian, Hua Wu, Haifeng Wang} \\
  Baidu Inc., China \\
  {\tt \{ouyangxuan, wangshuohuan, pangchao04, sunyu02\}@baidu.com} \\
  {\tt \{tianhao, wu\_hua, wanghaifeng\}@baidu.com} \\}

\begin{document}
\maketitle
\begin{abstract}
Recent studies have demonstrated that pre-trained cross-lingual models achieve impressive performance in downstream cross-lingual tasks. This improvement benefits from learning a large amount of monolingual and parallel corpora. Although it is generally acknowledged that parallel corpora are critical for improving the model performance, existing methods are often constrained by the size of parallel corpora, especially for low-resource languages. In this paper, we propose \textsc{Ernie-M}, a new training method that encourages the model to align the representation of multiple languages with monolingual corpora, to overcome the constraint that the parallel corpus size places on the model performance. Our key insight is to integrate back-translation into the pre-training process. We generate pseudo-parallel sentence pairs on a monolingual corpus to enable the learning of semantic alignments between different languages, thereby enhancing the semantic modeling of cross-lingual models. Experimental results show that \textsc{Ernie-M} outperforms existing cross-lingual models and delivers new state-of-the-art results in various cross-lingual downstream tasks.\footnote{Code and models are available at \url{https://github.com/PaddlePaddle/ERNIE}}
\end{abstract}

\section{Introduction}
Recent studies have demonstrated that the pre-training of cross-lingual language models can significantly improve their performance in cross-lingual natural language processing tasks \cite{devlin2018bert,lample2019cross,conneau2019unsupervised,liu2020multilingual}. Existing pre-training methods include multilingual masked language modeling (MMLM; \citealt{devlin2018bert}) and translation language modeling (TLM; \citealt{lample2019cross}), of which the key point is to learn a shared language-invariant feature space among multiple languages. 
MMLM implicitly models the semantic representation of each language in a unified feature space by learning them separately.
TLM is an extension of MMLM that is trained with a parallel corpus and captures semantic alignment by learning a pair of parallel sentences simultaneously. This study shows that the use of parallel corpora can significantly improve the performance in downstream cross-lingual understanding and generation tasks. However, the sizes of parallel corpora are limited~\cite{tran2020cross}, restricting the performance of the cross-lingual language model.

To overcome the constraint of the parallel corpus size on the model performance, we propose \textsc{Ernie-M}, a novel cross-lingual pre-training method to learn semantic alignment among multiple languages on monolingual corpora. Specifically, we propose cross-attention masked language modeling (CAMLM) to improve the cross-lingual transferability of the model on parallel corpora, and it trains the model to predict the tokens of one language by using another language. Then, we utilize the transferability learned from parallel corpora to enhance multilingual representation. We propose back-translation masked language modeling (BTMLM) to train the model, and this helps the model to learn sentence alignment from monolingual corpora. In BTMLM, a part of the tokens in the input monolingual sentences is predicted into the tokens of another language. We then concatenate the predicted tokens and the input sentences as pseudo-parallel sentences to train the model. In this way, the model can learn sentence alignment with only monolingual corpora and overcome the constraint of the parallel corpus size while improving the model performance.

\textsc{Ernie-M} is implemented on the basis of XLM-R \cite{conneau2019unsupervised}, and we evaluate its performance on five widely used cross-lingual benchmarks: XNLI \cite{conneau2018xnli} for cross-lingual natural language inference, MLQA \cite{lewis2019mlqa} for cross-lingual question answering, CoNLL \cite{sang2003introduction} for named entity recognition, cross-lingual paraphrase adversaries from word scrambling (PAWS-X) \cite{hu2020xtreme} for cross-lingual paraphrase identification, and Tatoeba \cite{hu2020xtreme} for cross-lingual retrieval. The experimental results demonstrate that \textsc{Ernie-M} outperforms existing cross-lingual models and achieves new state-of-the-art (SoTA) results.

\section{Related Work}
\subsection{Multilingual Language Models}

Existing multilingual language models can be classified into two main categories: (1) discriminative models; (2) generative models.

In the first category, a multilingual bidirectional encoder representation from transformers (mBERT; \citealt{devlin2018bert}) is pre-trained using MMLM on a monolingual corpus, which learns a shared language-invariant feature space among multiple languages. The evaluation results show that the mBERT achieves significant performance in downstream tasks \cite{wu2019beto}. XLM \cite{lample2019cross} is extended on the basis of mBERT using TLM, which enables the model to learn cross-lingual token alignment from parallel corpora. XLM-R \cite{conneau2019unsupervised} demonstrates the effects of models when trained on a large-scale corpus. It used 2.5T data extracted from Common Crawl \cite{wenzek2019ccnet} that involves 100 languages for MMLM training. The results show that a large-scale training corpus can significantly improve the performance of the cross-lingual model. Unicoder \cite{huang2019unicoder} achieves gains on downstream tasks by employing a multi-task learning framework to learn cross-lingual semantic representations with monolingual and parallel corpora. ALM \cite{yang2020alternating} improves the model's transferability by enabling the model to learn cross-lingual code-switch sentences. \textsc{InfoXLM} \cite{chi2020infoxlm} adds a contrastive learning task for cross-lingual model training. \textsc{Hictl} \cite{wei2020learning} learns cross-lingual semantic representation from multiple facets (at word-levels and sentence-levels) to improve the performance of cross-lingual models. VECO \cite{luo2020veco} presents a variable encoder-decoder framework to unify the understanding and generation tasks and achieves significant improvement in both downstream tasks.

The second category includes MASS \cite{song2019mass}, mBART \cite{liu2020multilingual}, XNLG \cite{chi2020cross} and mT5 \cite{xue2020mt5}. MASS \cite{vaswani2017attention} proposed a training objective for restore the input sentences in which successive token fragments are masked which improved the model's performance on machine translation. Similar to MASS, mBART pre-trains a denoised sequence-to-sequence model and uses an autoregressive task to train the model. XNLG focuses on multilingual question generation and abstractive summarization and updates the parameters of the encoder and decoder through auto-encoding and autoregressive tasks. mT5 uses the same model structure and pre-training method as T5 \cite{raffel2019exploring}, and extends the parameters of the cross-lingual model to 13B, significantly improving the performance of the cross-language downstream tasks.

\subsection{Back Translation and Non-Autoregressive Neural Machine Translation}
Back translation (BT) is an effective neural-network-based machine translation method proposed by \citet{sennrich2015improving}. It can significantly improve the performance of both supervised and unsupervised machine translation via augment the parallel training corpus \cite{lample2017unsupervised, edunov2018understanding}. BT has been found to particularly useful when the parallel corpus is sparse \cite{karakanta2018neural}. Predicting the token of the target language in one batch can also improve the speed of  non-auto regressive machine translation (NAT; \citealt{gu2017non,wang2019non}). Our work is inspired by NAT and BT. We generate the tokens of another language in batches and then use these in pre-training to help sentence alignment learning. 

\section{Methodology}
% \label{sec:length}
In this section, we first introduce the general workflow of \textsc{Ernie-M} and then present the details of the model training.

\begin{figure*}[!t]
\vskip 0.1in
\centering
\subfigure[MMLM]{\includegraphics[width=5cm]{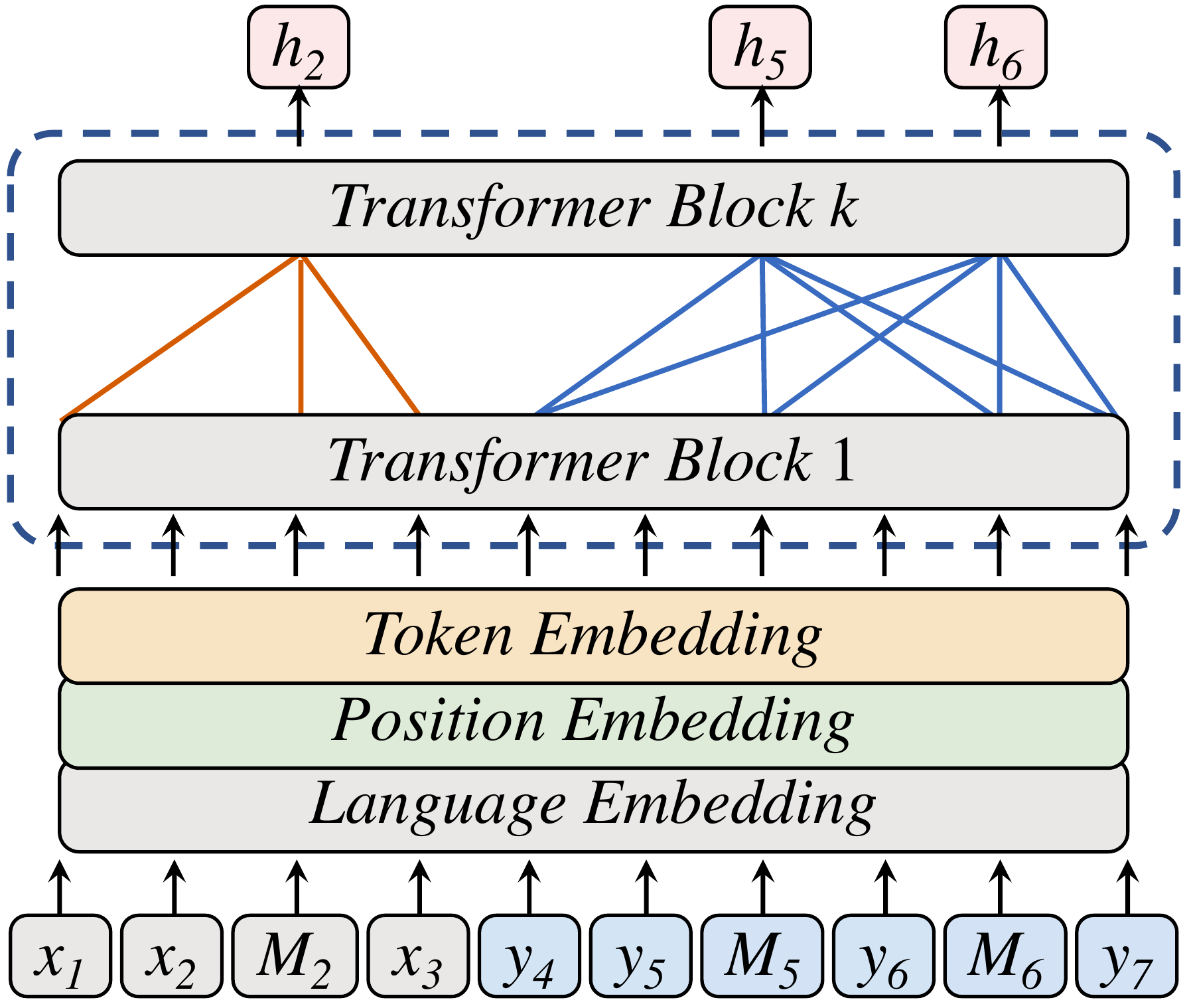}}
\subfigure[TLM]{\includegraphics[width=5cm]{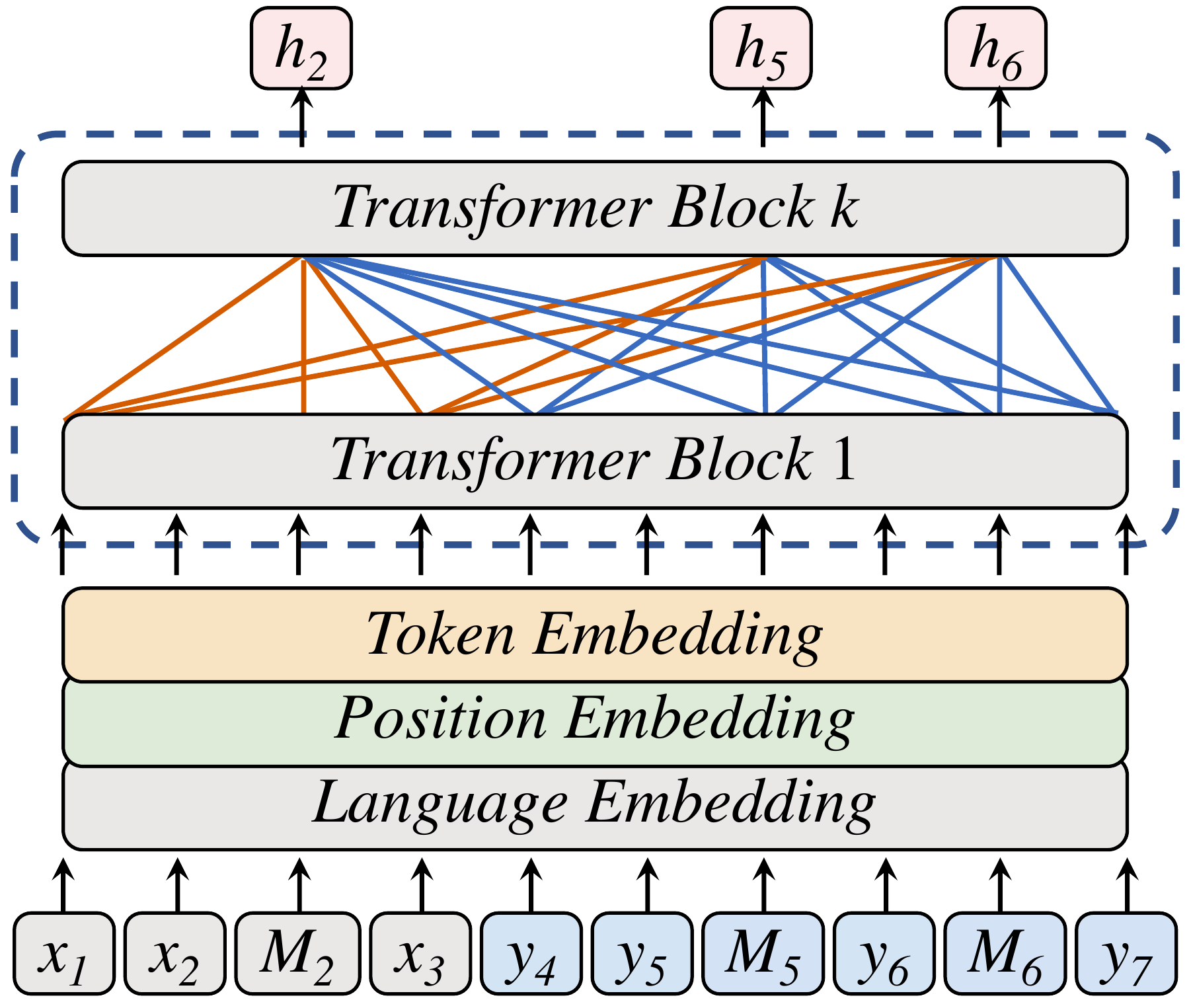}}
\subfigure[CAMLM]{\includegraphics[width=5cm]{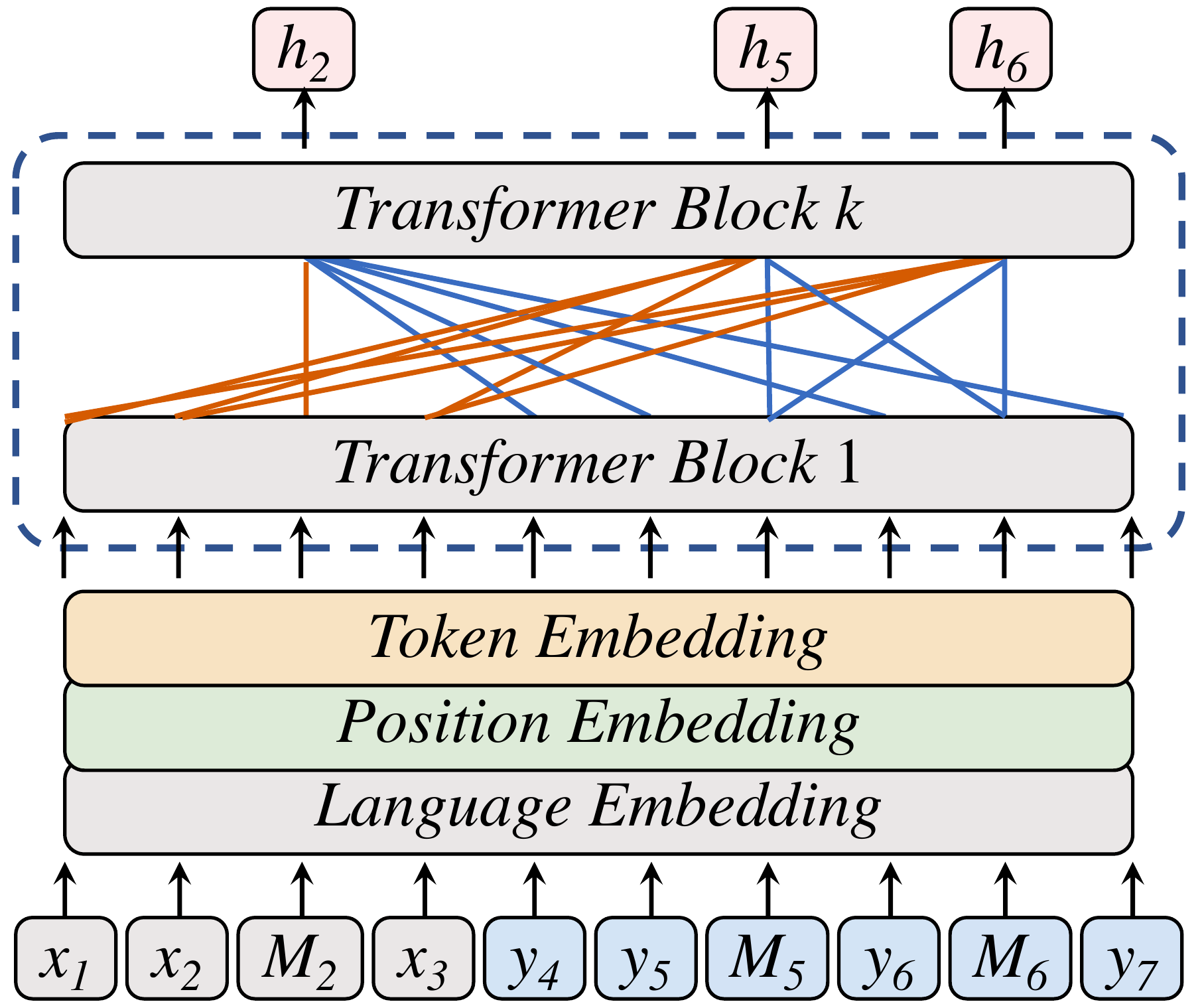}}
\caption{Overview of MMLM, TLM and CAMLM training. The input sentences in sub-figure (a) are monolingual sentences; $x$ and $y$ represent monolingual input sentences in different languages. The input sentences in sub-figures (b) and (c) are parallel sentences; $x$ and $y$ denote the source and target sentences of the parallel sentences, respectively. $h$  indicates the token predicted by the model.}
\vspace{-0.15in}
\label{fig1}
%\vskip -0.1in
\end{figure*}

% \subsection{Cross-lingual Semantic Alignment}
\paragraph{Cross-lingual Semantic Alignment.}
The key idea of \textsc{Ernie-M} is to utilize the transferability learned from parallel corpora to enhance the model's learning of large-scale monolingual corpora, and thus enhance the multilingual semantic representation. Based on this idea, we propose two pre-training objectives, cross-attention masked language modeling (CAMLM) and back-translation masked language modeling (BTMLM). 
CAMLM is to align the cross-lingual semantic representation on parallel corpora. 
Then, the transferability learned from parallel corpora is utilized to enhance the multilingual representation. Specifically, we train the \textsc{Ernie-M} by using BTMLM, enabling the model to align the semantics of multiple languages from monolingual corpora and improve the multilingual representation of the model. 
The MMLM and TLM are used by default because of the strong performance shown in \citealt{lample2019cross}. 
We combine MMLM, TLM with CAMLM, BTMLM to train \textsc{Ernie-M}. In the following sections, we will introduce the details of each objective.

\paragraph{Cross-attention Masked Language Modeling.} To learn the alignment of cross-lingual semantic representations in parallel corpora, we propose a new pre-training objective, CAMLM.
We denote a parallel sentence pair as $<$source sentence, target sentence$>$. In CAMLM, we learn the multilingual semantic representation by restoring the \texttt{MASK} token in the input sentences. When the model restores the \texttt{MASK} token in the source sentence, the model can only rely on the semantics of the target sentence, which means that the model has to learn how to represent the source language with the semantics of the target sentence and thus align the semantics of multiple languages.

\begin{figure}[!htp]
\centering
\vspace{-0.1in}
\subfigure[MMLM]{\includegraphics[width=2.6cm]{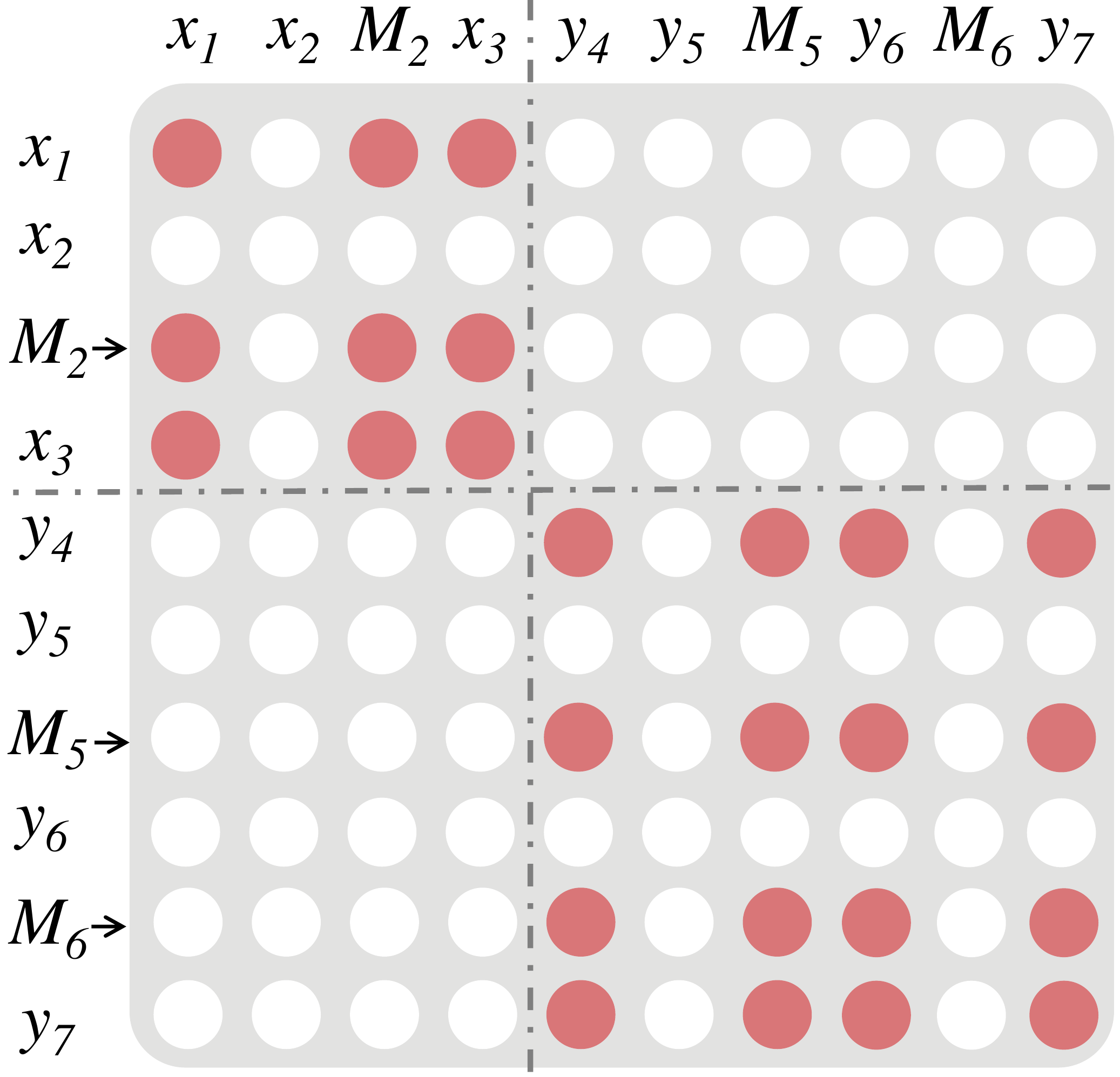}}
\subfigure[TLM]{\includegraphics[width=2.32cm]{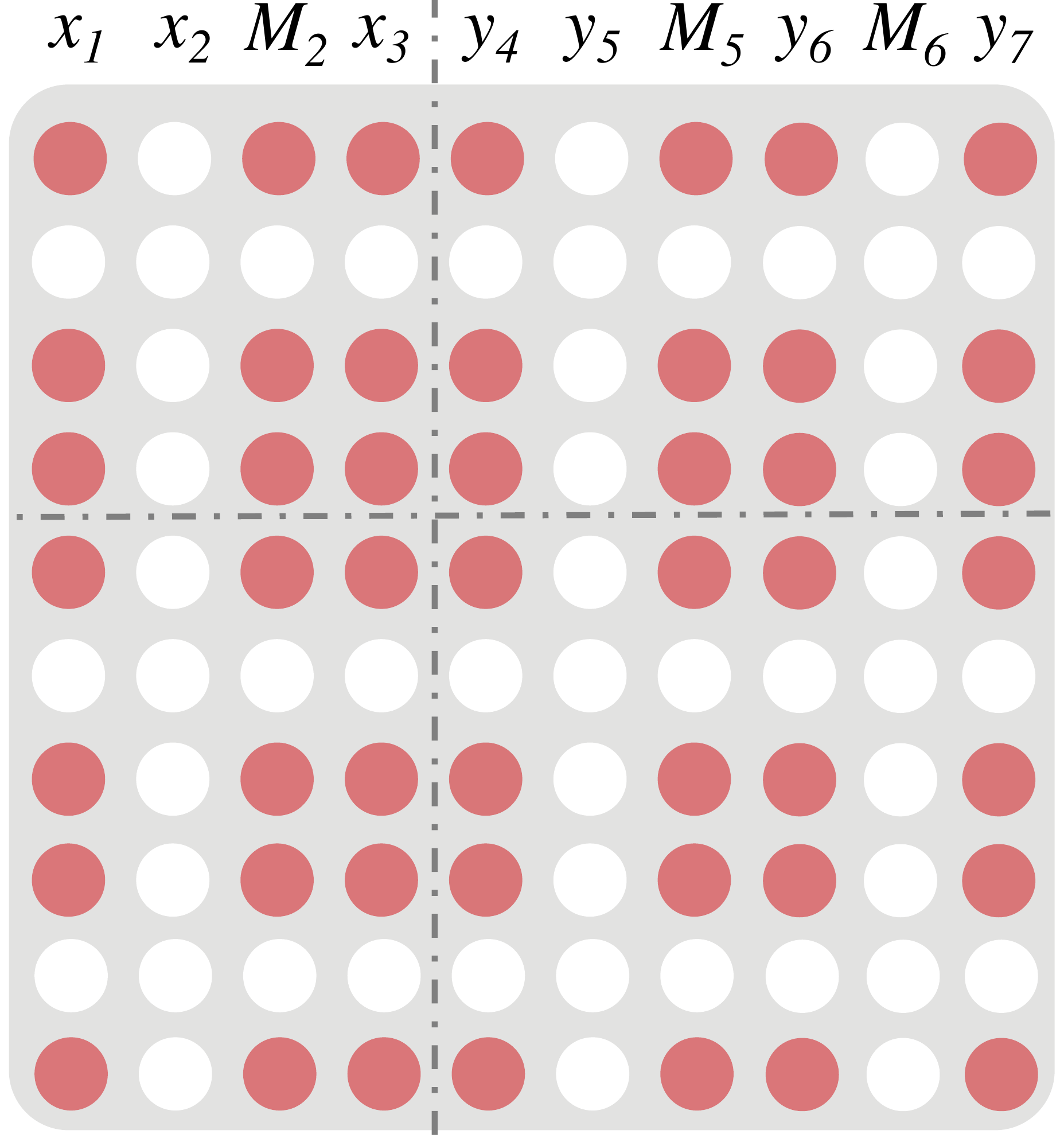}}
\subfigure[CAMLM]{\includegraphics[width=2.32cm]{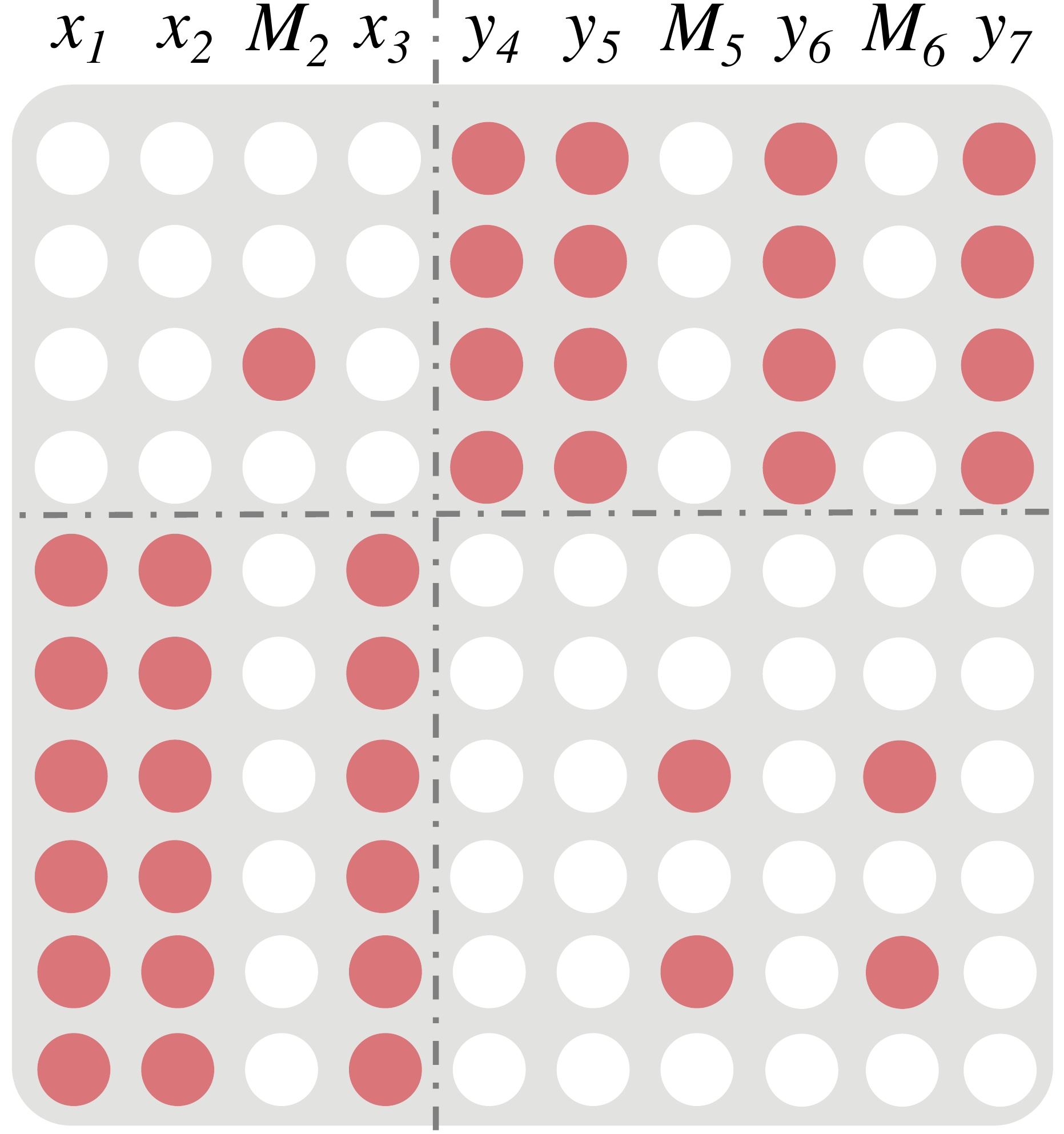}}
\caption{Self-attention mask matrix in MMLM, TLM and CAMLM. We use different self-attention masks for different pre-training objectives.}
\label{fig2}
\vskip -0in
\end{figure}

Figure \ref{fig1} (b) and (c) show the differences between TLM \cite{lample2019cross} and CAMLM. TLM learns the semantic alignment between languages with both the source and target sentences while CAMLM only relies on one side of the sentence to restore the \texttt{MASK} token. The advantage of CAMLM is that it avoids the information leakage that the model can attend to a  pair of input sentences at the same time, which makes learning of BTMLM  possible. The self-attention matrix of the example in Figure \ref{fig1} is shown in Figure \ref{fig2}. For TLM, the prediction of the \texttt{MASK} token relies on the input sentence pair. When the model learns CAMLM, the model can only predict the \texttt{MASK} token based on the sentence of its corresponding parallel sentence and the \texttt{MASK} symbol of this sentence, which provides the position and language information. Thus, the probability of the \texttt{MASK} token $M_{2}$ is $p(x_{2}|M_{2},y_{4},y_{5},y_{6},y_{7})$, $p(y_{5}|x_{1},x_{2},x_{3},M_{5})$ for $M_{5}$, and $p(y_{6}|x_{1},x_{2},x_{3},M_{6})$ for $M_{6}$ in CAMLM.

\begin{figure}[!ht]
    \centering
    \includegraphics[width=8cm]{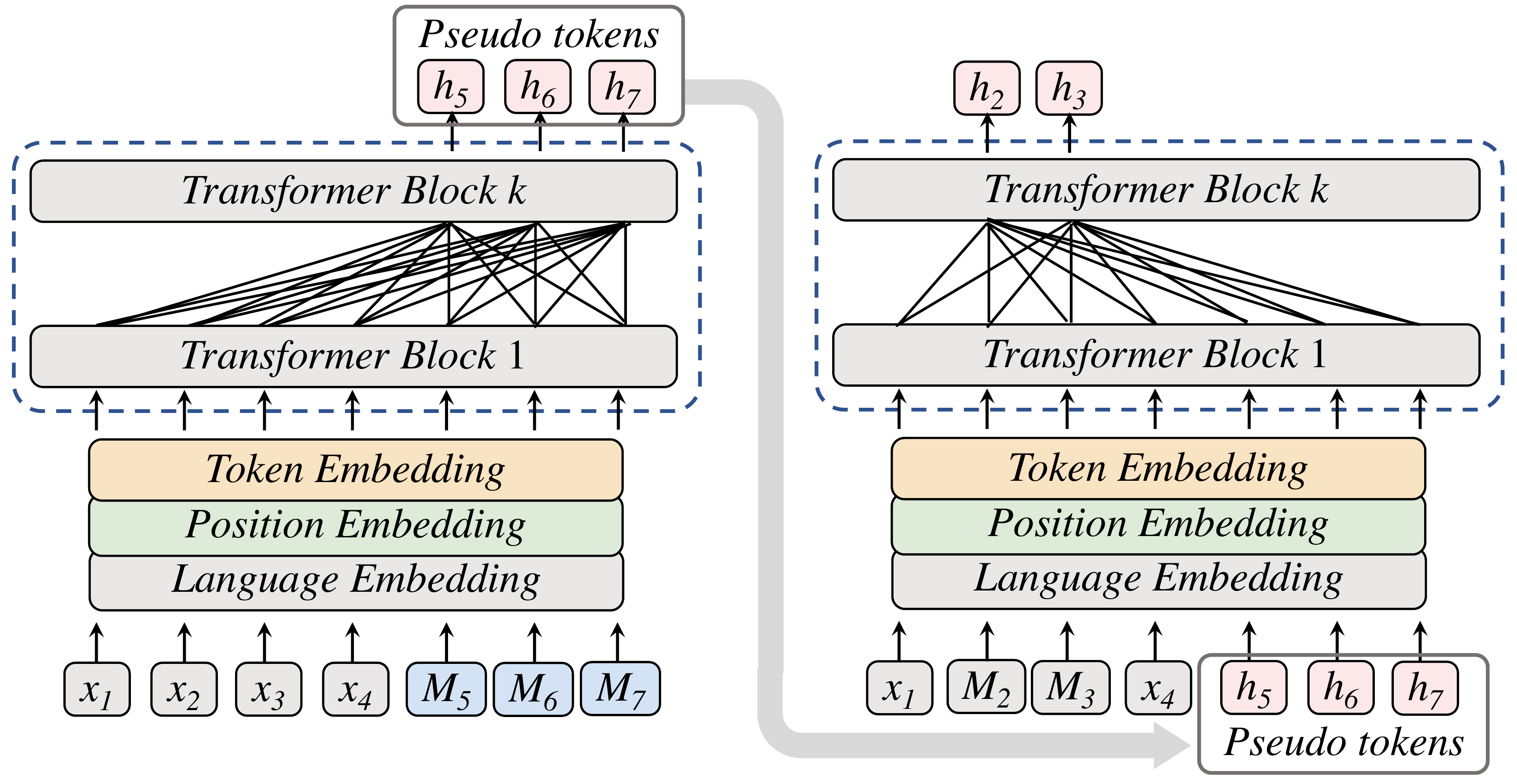}
    \vspace{-0.1in}
    \caption{Overview of BTMLM training; the left figure represents the first stage of BTMLM, predicting the pseudo-tokens. The right figure represents the second stage of the BTMLM, making predictions based on the predicted pseudo-tokens and original sentences.}
    \label{fig3}
    \vspace{-0in}
\end{figure}

Given the input in a bilingual corpus $ X_{src}=\{x_{1},x_{2},\cdots,x_{s} \}$, and its corresponding \texttt{MASK} position, $ M_{src}=\{m_{1},m_{2},\cdots,m_{ms}\} $, the target sentence is $ X_{tgt}=\{x_{s+1},x_{s+2},\cdots,x_{s+t}\}$, and its corresponding \texttt{MASK} position is $ M_{tgt}=\{m_{ms+1},m_{ms+2},\cdots,m_{ms+mt}\}$. In TLM, the model can attend to the tokens in the source and target sentences, so the probability of masked tokens is $ \prod_{m\in M} p(x_{m}|X/_{M})$, where $M=M_{src}\cup M_{tgt}$. $X/_{M}$ denotes all input tokens $x$ in $X$ except $x$ in $M$, where $X=X_{src}\cup X_{tgt}$. $x_{m}$ denotes the token with position $m$. In CAMLM, the probability of the \texttt{MASK} token in the source sentence is $ \prod_{m\in M_{src}} p(x_{m}|X/_{M\cup X_{src}})$, which means that when predicting the \texttt{MASK} tokens in the source sentence, we only focus on the target sentence. As for the target sentence, the probability of the \texttt{MASK} token is $ \prod_{m\in M_{tgt}} p(x_{m}|X/_{M\cup X_{tgt}})$, which means that the \texttt{MASK} tokens in the target sentence will be predicted based only on the source sentence. Therefore, the model must learn to use the corresponding sentence to predict and learn the alignment across multiple languages.
The pre-training loss of CAMLM in the source/target sentence is

$$
\mathcal{L}_{CAMLM(src)}=-\!\!\sum_{x\in D_{B}}\!\!log\!\!\prod_{m\in M_{src}}\!\! p(x_{m}|X/_{M\cup X_{src}})
$$
$$
\mathcal{L}_{CAMLM(tgt)}=-\!\!\sum_{x\in D_{B}}\!\!log\!\!\prod_{m\in M_{tgt}}\!\! p(x_{m}|X/_{M\cup X_{tgt}})
$$
where $D_{B}$ is the bilingual training corpus. The CAMLM loss is
$$
\mathcal{L}_{CAMLM}=\mathcal{L}_{CAMLM(src)}+\mathcal{L}_{CAMLM(tgt)}
$$
% \subsection{BTMLM}
\paragraph{Back-translation Masked Language Modeling.}
To overcome the constraint that the parallel corpus size places on the model performance, we propose a novel pre-training objective inspired by NAT \cite{gu2017non,wang2019non} and BT methods called BTMLM to align cross-lingual semantics with the monolingual corpus. We use BTMLM to train our model, which builds on the transferability learned through CAMLM, generating pseudo-parallel sentences from the monolingual sentences and the generated pseudo-parallel sentences are then used as the input of the model to align the cross-lingual semantics, thus enhancing the multilingual representation. The training process for BTMLM is shown in Figure \ref{fig3}.

The learning process for the BTMLM is divided into two stages. Stage 1 involves the generation of pseudo-parallel tokens from monolingual corpora. Specifically, we fill in several placeholder \texttt{MASK} at the end of the monolingual sentence to indicate the location and the language we want to generate, and let the model generate its corresponding parallel language token based on the original monolingual sentence and the corresponding position of the pseudo-token. In this way, we generate the tokens of another language from the monolingual sentence, which will be used in learning cross-lingual semantic alignment for multiple languages.

\begin{figure}[htp]
\vskip 0.1in
\centering
\includegraphics[width=4cm]{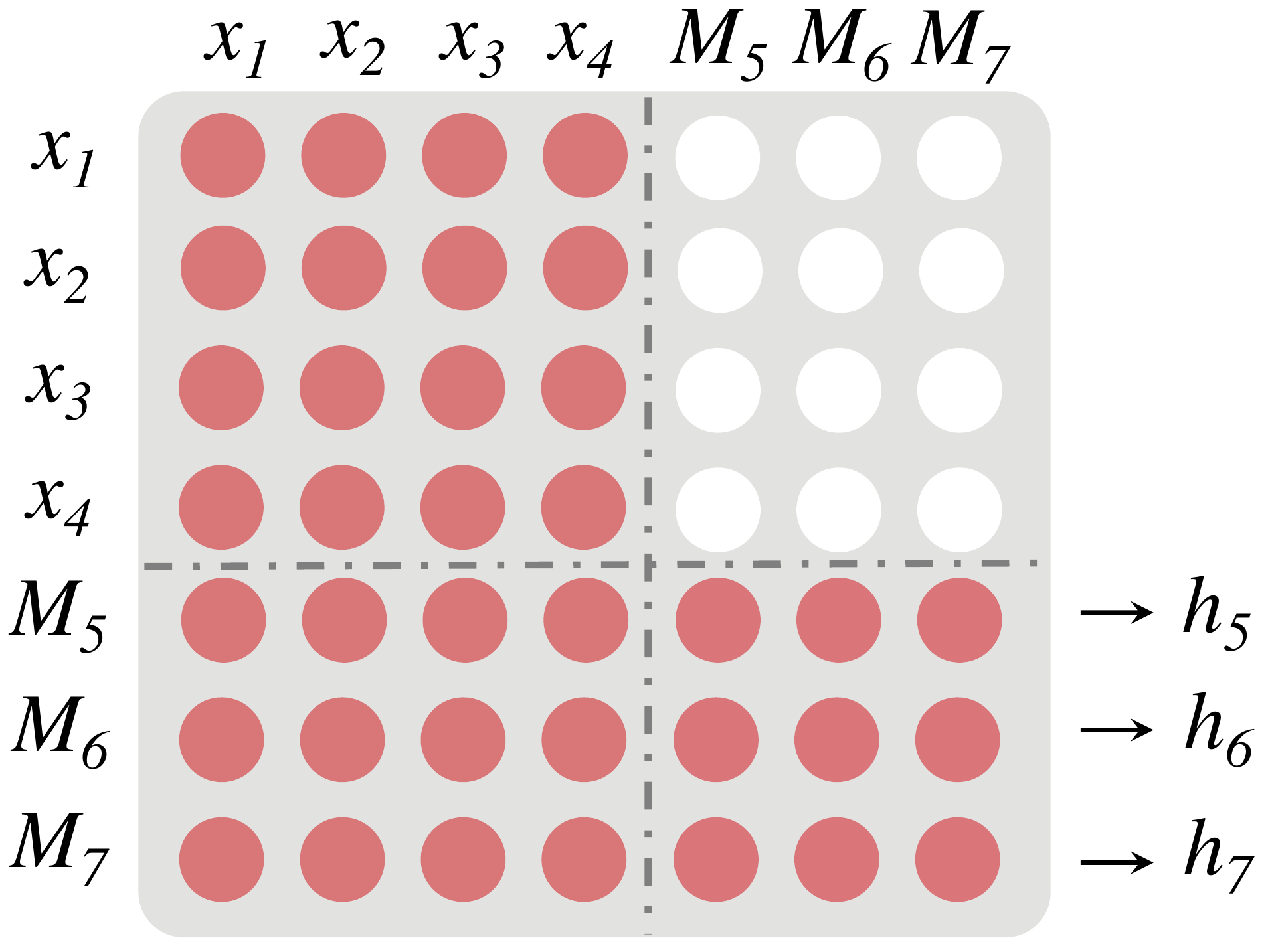}
\caption{Self-attention matrix of BTMLM Stage 1.}
\label{fig4}
\vskip -0in
\end{figure}

The self-attention matrix for generating pseudo-tokens in Figure \ref{fig3} is shown in Figure \ref{fig4}. In the pseudo-token generating process, the model can only attend to the source sentence and the placeholder \texttt{MASK} tokens, which indicate the language and position we want to predict by using language embedding and position embedding. The probability of mask token $M_{5}$ is $p(y_{5}|x_{1},x_{2},x_{3},x_{4}, M_{5})$, $p(y_{6}|x_{1},x_{2},x_{3},x_{4}, M_{6})$ for $M_{6}$ and  $p(y_{7}|x_{1},x_{2},x_{3},x_{4}, M_{7})$ for $M_{7}$.

Stage 2 uses the pseudo-tokens generated in Stage 1 to learn the cross-lingual semantics alignment. The process in Stage 2 is shown in the right-hand diagram of Figure \ref{fig3}. In the training process of Stage 2, the input of the model is the concatenation of the monolingual sentences and the generated pseudo-parallel tokens, and the learning objective is to restore the \texttt{MASK} tokens based on the original sentences and the generated pseudo-parallel tokens. Because the model can rely not only on the input monolingual sentence but also the generated pseudo-tokens in the process of inference \texttt{MASK} tokens, the model can explicitly learn the alignment of the cross-lingual semantic representation from the monolingual sentences.

The learning process of the BTMLM can be interpreted as follows: given the input in monolingual corpora $X=\{x_{1},x_{2},\cdots,x_{s}\}$, the positions of masked tokens $ M=\{m_{1},m_{2},\cdots,m_{m}\}$ and the position of the pseudo-token to be predicted, $M_{pseudo}=\{m_{s+1},m_{s+2},\cdots,m_{s+p}\}$, we first generate pseudo-tokens $P=\{h_{s+1},h_{s+2},\cdots,h_{s+p}\}$, as described earlier; we then concatenate the generated pseudo-token with input monolingual sentence as a new parallel sentence pair and use it to train our model. Thus, the probability of the masked tokens in BTMLM is $\prod_{m\in M} p(x_{m}|X/_{M},P)$, where $X/_{M}$ denotes all input tokens $x$ in $X$ except $x$ in $M$. The pre-training loss of BTMLM is 

$$
\mathcal{L}_{BTMLM}=-\sum_{x\in D_{M}}log\prod_{m\in M} p(x_{m}|X/_{M}, P)
$$
where $D_{M}$ is the monolingual training corpus. 

%\begin{table*}[!htbp]
\renewcommand\arraystretch{0.9}

\begin{table*}[!h]
\centering
\vskip 0.1in

\resizebox{\textwidth}{!}{

\begin{tabular}{l|ccccccccccccccc|c}
\toprule
\textbf{Model}& \textbf{en}& \textbf{fr}& \textbf{es}& \textbf{de}& \textbf{el}& \textbf{bg}& \textbf{ru}& \textbf{tr}& \textbf{ar}& \textbf{vi}& \textbf{th}& \textbf{zh}& \textbf{hi}& \textbf{sw}& \textbf{ur}& \textbf{Avg}  \\
\midrule
\multicolumn{17}{l}{\textit{Fine-tune cross-lingual model on English training set (Cross-lingual Transfer)}} \\
\midrule
XLM \cite{lample2019cross} & 85.0 & 78.7 & 78.9 & 77.8 & 76.6 & 77.4 & 75.3 & 72.5 & 73.1 & 76.1 & 73.2 & 76.5 & 69.6 & 68.4 & 67.3 & 75.1 \\
Unicoder \cite{huang2019unicoder} & 85.1 & 79.0 & 79.4 & 77.8 & 77.2 & 77.2 & 76.3 & 72.8 & 73.5 & 76.4 & 73.6 & 76.2 & 69.4 & 69.7 & 66.7 & 75.4 \\
XLM-R \cite{conneau2019unsupervised} & 85.8 & 79.7 & 80.7 & 78.7 & 77.5 & 79.6 & 78.1 & 74.2 & 73.8 & 76.5 & 74.6 & 76.7 & 72.4 & 66.5 & 68.3 & 76.2 \\
\textsc{InfoXLM} \cite{chi2020infoxlm} & \textbf{86.4} & \textbf{80.6} & 80.8 & 78.9 & 77.8 & 78.9 & 77.6 & 75.6 & 74.0 & 77.0 & 73.7 & 76.7 & 72.0 & 66.4 & 67.1 & 76.2 \\
\textsc{Ernie-M} & 85.5 & 80.1 & \textbf{81.2} & \textbf{79.2} & \textbf{79.1} & \textbf{80.4} & \textbf{78.1} & \textbf{76.8} & \textbf{76.3} & \textbf{78.3} & \textbf{75.8} & \textbf{77.4} & \textbf{72.9} & \textbf{69.5} & \textbf{68.8} & \textbf{77.3} \\

\midrule
XLM-R$_{\scriptsize \textsc{Large}}$ \cite{conneau2019unsupervised} & 89.1 & 84.1 & 85.1 & 83.9 & 82.9 & 84.0 & 81.2 & 79.6 & 79.8 & 80.8 & 78.1 & 80.2 & 76.9 & 73.9 & 73.8 & 80.9 \\
$\textsc{InfoXLM}_{\scriptsize \textsc{Large}}$ \cite{chi2020infoxlm} & \textbf{89.7} & 84.5 & 85.5 & 84.1 & 83.4 & 84.2 & 81.3 & 80.9 & 80.4 & 80.8 & 78.9 & 80.9 & 77.9 & 74.8 & 73.7 & 81.4 \\
VECO$_{\scriptsize \textsc{Large}}$ \cite{luo2020veco} & 88.2 & 79.2 & 83.1 & 82.9 & 81.2 & 84.2 & \textbf{82.8} & 76.2 & 80.3 & 74.3 & 77.0 & 78.4 & 71.3 & \textbf{80.4} & \textbf{79.1} & 79.9 \\
$\textsc{Ernie-M}_{\scriptsize \textsc{Large}}$ & 89.3 & \textbf{85.1} & \textbf{85.7} & \textbf{84.4} & \textbf{83.7} & \textbf{84.5} & 82.0 & \textbf{81.2} & \textbf{81.2} & \textbf{81.9} & \textbf{79.2} & \textbf{81.0} & \textbf{78.6} & 76.2 & 75.4 & \textbf{82.0} \\

\midrule
\multicolumn{17}{l}{\textit{Fine-tune cross-lingual model on all training sets (Translate-Train-All)}} \\
\midrule
XLM \cite{lample2019cross}& 85.0 & 80.8 & 81.3 & 80.3 & 79.1 & 80.9 & 78.3 & 75.6 & 77.6 & 78.5 & 76.0 & 79.5 & 72.9 & 72.8 & 68.5 & 77.8 \\
Unicoder \cite{huang2019unicoder}& 85.6 & 81.1 & 82.3 & 80.9 & 79.5 & 81.4 & 79.7 & 76.8 & 78.2 & 77.9 & 77.1 & 80.5 & 73.4 & 73.8 & 69.6 & 78.5 \\
XLM-R \cite{conneau2019unsupervised}& 85.4 & 81.4 & 82.2 & 80.3 & 80.4 & 81.3 & 79.7 & 78.6 & 77.3 & 79.7 & 77.9 & 80.2 & 76.1 & 73.1 & 73.0 & 79.1 \\
\textsc{InfoXLM} \cite{chi2020infoxlm}& 86.1 & 82.0 & 82.8 & 81.8 & 80.9 & 82.0 & 80.2 & 79.0 & 78.8 & 80.5 & 78.3 & 80.5 & 77.4 & 73.0 & 71.6 & 79.7 \\
\textsc{Ernie-M} & \textbf{86.2} & \textbf{82.5} & \textbf{83.8} & \textbf{82.6} & \textbf{82.4} & \textbf{83.4} & \textbf{80.2} & \textbf{80.6} & \textbf{80.5} & \textbf{81.1} & \textbf{79.2} & \textbf{80.5} & \textbf{77.7} & \textbf{75.0} & \textbf{73.3} & \textbf{80.6} \\
\midrule
XLM-R$_{\scriptsize \textsc{Large}}$ \cite{conneau2019unsupervised}& 89.1 & 85.1 & 86.6 & 85.7 & 85.3 & 85.9 & 83.5 & 83.2 & 83.1 & 83.7 & 81.5 & \textbf{83.7} & \textbf{81.6} & 78.0 & 78.1 & 83.6 \\
VECO$_{\scriptsize \textsc{Large}}$ \cite{luo2020veco}& 88.9 & 82.4 & 86.0 & 84.7 & 85.3 & 86.2 & \textbf{85.8} & 80.1 & 83.0 & 77.2 & 80.9 & 82.8 & 75.3 & \textbf{83.1} & \textbf{83.0} & 83.0 \\
$\textsc{Ernie-M}_{\scriptsize \textsc{Large}}$ & \textbf{89.5} & \textbf{86.5} & \textbf{86.9} & \textbf{86.1} & \textbf{86.0} & \textbf{86.8} & 84.1 & \textbf{83.8} & \textbf{84.1} & \textbf{84.5} & \textbf{82.1} & 83.5 & 81.1 & 79.4 & 77.9 & \textbf{84.2} \\
\bottomrule

\end{tabular}}
\caption{Evaluation results on XNLI cross-lingual natural language inference. We report the accuracy on each of the 15 XNLI languages and the average accuracy. Our \textsc{Ernie-M} results are based on five runs with different random seeds.}
\label{table1}
\vskip -0.1in
\end{table*}

\section{Experiments}
We consider five cross-lingual evaluation benchmarks: XNLI for cross-lingual natural language inference, MLQA for cross-lingual question answering, CoNLL for cross-lingual named entity recognition, PAWS-X for cross-lingual paraphrase identification, and Tatoeba for cross-lingual retrieval. Next, we first describe the data and pre-training details and then compare the \textsc{Ernie-M} with the existing state-of-the-art models.

\subsection{Data and Model}
\textsc{Ernie-M} is trained with monolingual and parallel corpora that involved 96 languages. For the monolingual corpus, we extract it from CC-100 according to \citet{wenzek2019ccnet, conneau2019unsupervised}. For the bilingual corpus, we use the same corpus as \textsc{InfoXLM} \cite{chi2020infoxlm}, including MultiUN \cite{ziemski2016united}, IIT Bombay \cite{kunchukuttan2017iit}, OPUS \cite{tiedemann2012parallel}, and WikiMatrix \cite{schwenk2019wikimatrix}

We use a transformer-encoder \cite{vaswani2017attention} as the backbone of the model. For the $\textsc{Ernie-M}_{\scriptsize \textsc{Base}}$ model, we adopt a structure with 12 layers, 768 hidden units, 12 heads. For $\textsc{Ernie-M}_{\scriptsize \textsc{Large}}$ model , we adopt a structure with 24 layers, 1024 hidden units, 16 heads. The activation function used is GeLU \cite{hendrycks2016gaussian}. Following \citealt{chi2020infoxlm} and \citealt{luo2020veco}, we initialize the parameters of \textsc{Ernie-M} with XLM-R. We use the Adam optimizer \cite{kingma2014adam} to train \textsc{Ernie-M}; the learning rate is scheduled with a linear decay with 10K warm-up steps, and the peak learning rate is $2e-4$ for the base model and $1e-4$ for the large model. We conduct the pre-training experiments using 64 Nvidia V100-32GB GPUs with 2048 batch size and 512 max length. 

\begin{table}[!t]
\centering

\vskip 0.1in

\scalebox{0.8}{
\begin{tabular}{l|cccc|c}
\toprule
\textbf{Model}& \textbf{en}& \textbf{nl}& \textbf{es}& \textbf{de}& \textbf{Avg}  \\
\midrule
\multicolumn{6}{l}{\textit{Fine-tune on English dataset}} \\
\midrule
mBERT$^*$ & 91.97 & 77.57 & 74.96 & 69.56 & 78.52 \\
XLM-R$^\dag$ & 92.25 & \textbf{78.08} & 76.53 & \textbf{69.60} & 79.11 \\
\textsc{Ernie-M} & \textbf{92.78} & 78.01 & \textbf{79.37} & 68.08 & \textbf{79.56} \\
\midrule
XLM-R$^\dag_{\scriptsize \textsc{Large}}$ & 92.92 & 80.80 & 78.64 & 71.40 & 80.94 \\
\textsc{Ernie-M}$_{\scriptsize \textsc{Large}}$ & \textbf{93.28} & \textbf{81.45} & \textbf{78.83} & \textbf{72.99} & \textbf{81.64} \\
\midrule
\multicolumn{6}{l}{\textit{Fine-tune on all dataset}} \\
\midrule
XLM-R$^\dag$ & 91.08 & 89.09 & 87.28 & 83.17 & 87.66  \\
\textsc{Ernie-M} & \textbf{93.04} & \textbf{91.73} & \textbf{88.33} & \textbf{84.20} & \textbf{89.32} \\
\midrule
XLM-R$^\dag_{\scriptsize \textsc{Large}}$ & 92.00 & 91.60 & \textbf{89.52} & 84.60 & 89.43  \\
\textsc{Ernie-M}$_{\scriptsize \textsc{Large}}$ & \textbf{94.01} & \textbf{93.81} & 89.23 & \textbf{86.20} & \textbf{90.81} \\
\bottomrule
\end{tabular}}
\caption{Evaluation results on CoNLL named entity recognition. The results of \textsc{Ernie-M} are averaged over five runs. Results with ``$\dag$'' and ``$*$'' are from \cite{conneau2019unsupervised}, and \cite{wu2019beto}, respectively.}
\label{table2}
\vskip -0.2in
\end{table}

\begin{table*}[!htbp]
\centering

\vskip -0.5in

\resizebox{\textwidth}{!}{
\begin{tabular}{l|ccccccc|c}
\toprule
\textbf{Model}& \textbf{en}& \textbf{es}& \textbf{de}& \textbf{ar}& \textbf{hi}& \textbf{vi}& \textbf{zh}& \textbf{Avg}  \\
\midrule

mBERT \cite{lewis2019mlqa}& 77.7 / 65.2 &  64.3 / 46.6 & 57.9 / 44.3 & 45.7 / 29.8 & 43.8 / 29.7 & 57.1 / 38.6 & 57.5 / 37.3 & 57.7 / 41.6 \\
XLM \cite{lewis2019mlqa}& 74.9 / 62.4 &  68.0 / 49.8 & 62.2 / 47.6 & 54.8 / 36.3 & 48.8 / 27.3 & 61.4 / 41.8 & 61.1 / 39.6 & 61.6 / 43.5 \\
XLM-R \cite{conneau2019unsupervised}& 77.1 / 64.6 &  67.4 / 49.6 & 60.9 / 46.7 & 54.9 / 36.6 & 59.4 / 42.9 & 64.5 / 44.7 & 61.8 / 39.3 & 63.7 / 46.3 \\
\textsc{InfoXLM} \cite{chi2020infoxlm}& 81.3 / 68.2 &  69.9 / 51.9 & 64.2 / 49.6 & 60.1 / 40.9 & 65.0 / 47.5 & 70.0 / 48.6 & 64.7 / \textbf{41.2} & 67.9 / 49.7 \\
\textsc{Ernie-M} & \textbf{81.6} / \textbf{68.5} & \textbf{70.9} / \textbf{52.6} & \textbf{65.8} / \textbf{50.7} & \textbf{61.8} / \textbf{41.9} & \textbf{65.4} / \textbf{47.5} & \textbf{70.0} / \textbf{49.2} & \textbf{65.6} / 41.0 & \textbf{68.7} / \textbf{50.2} \\

\midrule
XLM-R$_{\scriptsize \textsc{Large}}$ \cite{conneau2019unsupervised}& 80.6 / 67.8 &  74.1 / 56.0 & 68.5 / 53.6 & 63.1 / 43.5 & 62.9 / 51.6 & 71.3 / 50.9 & 68.0 / 45.4 & 70.7 / 52.7 \\
\textsc{InfoXLM}$_{\scriptsize \textsc{Large}}$ \cite{chi2020infoxlm}& \textbf{84.5} / \textbf{71.6} &  \textbf{75.1} / \textbf{57.3} & \textbf{71.2} / \textbf{56.2} & \textbf{67.6} / \textbf{47.6} & 72.5 / 54.2 & \textbf{75.2} / \textbf{54.1} & 69.2 / 45.4 & 73.6 / 55.2 \\
\textsc{Ernie-M}$_{\scriptsize \textsc{Large}}$ & 84.4 / 71.5 & 74.8 / 56.6 & 70.8 / 55.9 & 67.4 / 47.2 & \textbf{72.6} / \textbf{54.7} & 75.0 / 53.7 & \textbf{71.1} / \textbf{47.5} & \textbf{73.7} / \textbf{55.3} \\

\bottomrule
\end{tabular}}
\caption{Evaluation results on MLQA cross-lingual question answering. We report the F1 and exact match (EM) scores. The results of \textsc{Ernie-M} are averaged over five runs.}
\label{table3}
\vskip -0.1in
\end{table*}

\subsection{Experimental Evaluation}
\paragraph{Cross-lingual Natural Language Inference.}
The cross-lingual natural language inference (XNLI; \citealt{conneau2018xnli}) task is a multilingual language inference task. The goal of XNLI is to determine the relationship between the two input sentences. We evaluate \textsc{Ernie-M} in (1) cross-lingual transfer \cite{conneau2018xnli} setting: fine-tune the model with an English training set and evaluate the foreign language XNLI test and (2) translate-train-all \cite{huang2019unicoder} setting: fine-tune the model on the concatenation of all other languages and evaluate on each language test set.

Table \ref{table1} shows the results of \textsc{Ernie-M} in XNLI task. The result shows that \textsc{Ernie-M} outperforms all baseline models including XLM \cite{lample2019cross}, Unicoder \cite{huang2019unicoder}, XLM-R \cite{conneau2019unsupervised}, \textsc{InfoXLM} \cite{chi2020infoxlm} and VECO \cite{luo2020veco} on both the evaluation settings on XNLI. The final scores on the test set are averaged over five runs with different random seeds. On cross-lingual transfer setting, \textsc{Ernie-M} achieves 77.3 average accuracy, outperforming \textsc{InfoXLM} by 1.1, \textsc{Ernie-M}$_{\scriptsize \textsc{Large}}$ achieves 82.0 accuracy, outperforming \textsc{InfoXLM}$_{\scriptsize \textsc{Large}}$ by 0.6. \textsc{Ernie-M} also yields outstanding performance in low-resource languages, including 69.5 in Swahili (sw) and 68.8 in Urdu (ur). In the case of translate-train-all, \textsc{Ernie-M} improves the performance and reaches an accuracy of 80.6, outperforming \textsc{InfoXLM} by 0.9, \textsc{Ernie-M}$_{\scriptsize \textsc{Large}}$ achieves 84.2 accuracy, a new SoTA for XNLI, outperforming XLM-R$_{\scriptsize \textsc{Large}}$ by 0.6.

\paragraph{Named Entity Recognition.}
For the named-entity-recognition task, we evaluate \textsc{Ernie-M} on the CoNLL-2002 and CoNLL-2003 datasets \cite{sang2003introduction}, which is a cross-lingual named-entity-recognition task including English, Dutch, Spanish and German. We consider \textsc{Ernie-M} in the following setting: (1) fine-tune on the English dataset and evaluate on each cross-lingual dataset to evaluate cross-lingual transfer and (2) fine-tune on all training datasets to evaluate cross-lingual learning. For each setting, we reported the F1 score for each language.

Table \ref{table2} shows the results of \textsc{Ernie-M}, XLM-R, and mBERT on CoNLL-2002 and CoNLL-2003. The results of XLM-R and mBERT are reported from \citet{conneau2019unsupervised}. \textsc{Ernie-M} model yields SoTA performance on both settings and outperforms XLM-R by 0.45 F1 when trained on English and 0.70 F1 when trained on all languages in the base model. Similar to the performance in the XNLI task, \textsc{Ernie-M} shows better performance on low-resource languages. For large models and fine-tune in all languages setting, \textsc{Ernie-M} is 2.21 F1 higher than SoTA in Dutch (nl) and 1.6 F1 higher than SoTA in German (de).

\paragraph{Cross-lingual Question Answering.} For the question answering task, we use a multilingual question answering (MLQA) dataset to evaluate \textsc{Ernie-M}. MLQA has the same format as SQuAD v1.1 \cite{rajpurkar2016squad} and is a multilingual language question answering task composed of seven languages. We fine-tune \textsc{Ernie-M} by training on English data and evaluating on seven cross-lingual datasets. The fine-tune method is the same as in \citet{lewis2019mlqa}, which concatenates the question-passage pair as the input.

Table \ref{table3} presents a comparison of \textsc{Ernie-M} and several baseline models on MLQA. We report the F1 and extract match (EM) scores based on the average over five runs. The performance of \textsc{Ernie-M} in MLQA is significantly better than the previous models, and it achieves a SoTA score. We outperform \textsc{InfoXLM} 0.8 in F1 and 0.5 in EM.

\paragraph{Cross-lingual Paraphrase Identification.} For cross-lingual paraphrase identification task, we use the PAWS-X \cite{hu2020xtreme} dataset to evaluate our model. The goal of PAWS-X was to determine whether two sentences were paraphrases. We evaluate \textsc{Ernie-M} on both the cross-lingual transfer setting and translate-train-all setting.

Table~\ref{table3.1} shows a comparison of \textsc{Ernie-M} and various baseline models on PAWS-X. We report the accuracy score on each language test set based on the average over five runs. The results show that \textsc{Ernie-M} outperforms all baseline models on most languages and achieves a new SoTA. 

\begin{table}[!t]
\centering

\vskip 0.1in

\scalebox{0.68}{
\begin{tabular}{l|ccccccc|c}
\toprule
\textbf{Model}& \textbf{en}& \textbf{de}& \textbf{es}& \textbf{fr}& \textbf{ja}&\textbf{ko}&\textbf{zh}&\textbf{Avg}  \\
\midrule
\multicolumn{9}{l}{\textit{Cross-lingual Transfer}} \\
\midrule
mBERT$^\dag$ & 94.0 & 85.7 & 87.4 & 87.0 & 73.0 & 69.6 & 77.0 & 81.9 \\
XLM$^\dag$ & 94.0 & 85.9 & 88.3 & 87.4 & 69.3 & 64.8 & 76.5 & 80.9 \\
MMTE$^\dag$ & 93.1 & 85.1 & 87.2 & 86.9 & 72.0 & 69.2 & 75.9 & 81.3 \\
XLM-R$^\dag_{\scriptsize \textsc{Large}}$ & 94.7 & 89.7 & 90.1 & 90.4 & 78.7 & 79.0 & 82.3 & 86.4 \\
VECO$^*_{\scriptsize \textsc{Large}}$ & \textbf{96.2} & 91.3 & 91.4 & 92.0 & 81.8 & 82.9 & 85.1 & 88.7 \\
\textsc{Ernie-M}$_{\scriptsize \textsc{Large}}$ & 96.0 & \textbf{91.9} & \textbf{91.4} & \textbf{92.2} & \textbf{83.9} & \textbf{84.5} & \textbf{86.9} & \textbf{89.5}\\
\midrule
\multicolumn{9}{l}{\textit{Translate-Train-All}} \\
\midrule
VECO$^*_{\scriptsize \textsc{Large}}$ & 96.4 & 93.0 & 93.0 & 93.5 & 87.2 & 86.8 & 87.9 & 91.1\\
\textsc{Ernie-M}$_{\scriptsize \textsc{Large}}$ & \textbf{96.5} & \textbf{93.5} & \textbf{93.3} & \textbf{93.8} & \textbf{87.9} & \textbf{88.4} & \textbf{89.2} & \textbf{91.8}\\
\bottomrule
\end{tabular}}
\caption{Evaluation results on PAWS-X. The results of \textsc{Ernie-M} are averaged over five runs. Results with ``$\dag$'' and ``$*$'' are from \cite{hu2020xtreme} and \cite{luo2020veco}, respectively.}
\label{table3.1}
\vskip -0.1in
\end{table}

\paragraph{Cross-lingual Sentence Retrieval.} The goal of the cross-lingual sentence retrieval task was to extract parallel sentences from bilingual corpora. We used a subset of the Tatoeba \cite{hu2020xtreme} dataset,  which contains 36 language pairs to evaluate \textsc{Ernie-M}. Following \citealt{luo2020veco}, we used the averaged representation in the middle layer of the best XNLI model to evaluate the retrieval task.

Table \ref{table3.2} shows the results of \textsc{Ernie-M} in the retrieval task; XLM-R results are reported from \citealt{luo2020veco}. \textsc{Ernie-M} achieves a score of 87.9 in the Tatoeba dataset, outperforming VECO 1.0 and obtaining new SoTA results. 

\begin{table}[!h]
\centering
\vskip 0.1in
%\resizebox{\textwidth}{!}{
\scalebox{0.7}{
\begin{tabular}{l|c}
\toprule
\textbf{Model}& \textbf{Avg} \\
\midrule
XLM-R$_{\scriptsize \textsc{Large}}$ \cite{luo2020veco}& 75.2  \\
VECO$_{\scriptsize \textsc{Large}}$ \cite{luo2020veco}& 86.9  \\
\textsc{Ernie-M}$_{\scriptsize \textsc{Large}}$ & \textbf{87.9}  \\
\midrule
\textsc{Ernie-M}$^\dag_{\scriptsize \textsc{Large}}$ & 93.3  \\
\bottomrule
\end{tabular}}
\caption{Evaluation results on Tatoeba. ``$\dag$'' indicates the results after fine-tuning.}
\label{table3.2}
\vskip -0 in
\end{table}

\begin{figure}[htp]
\centering
\includegraphics[width=8cm]{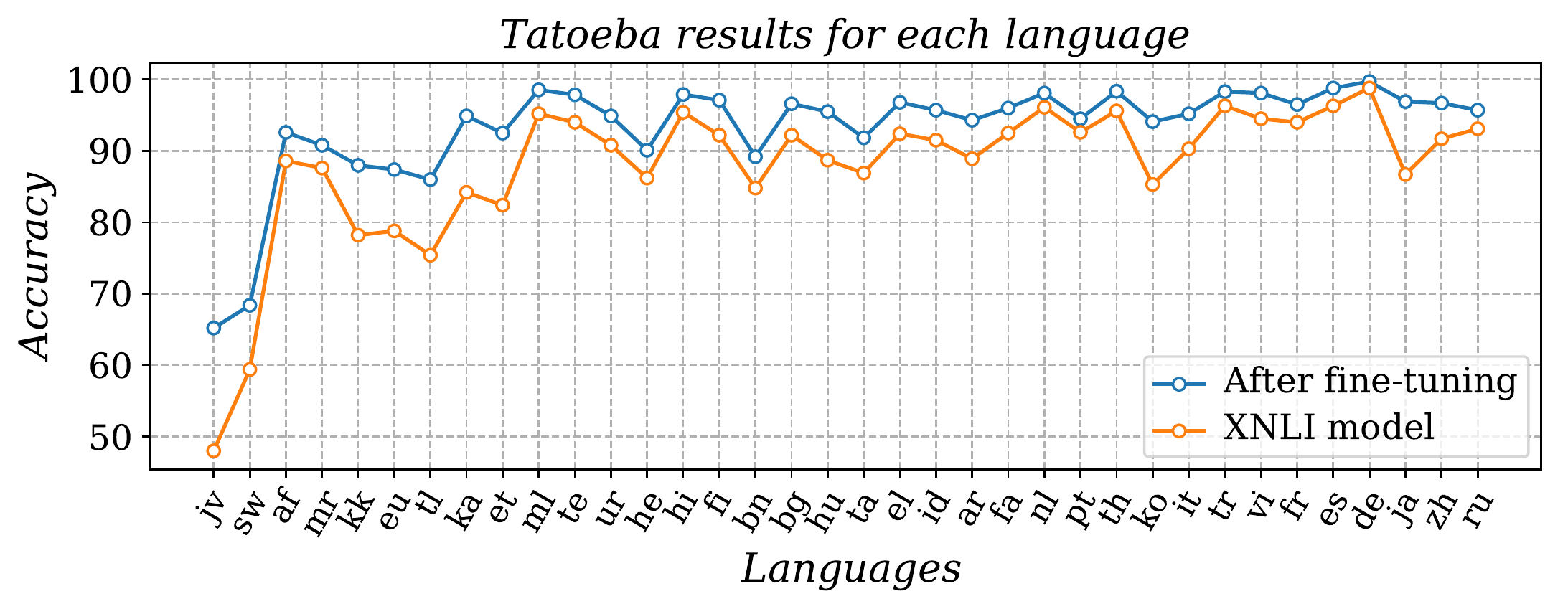}
\caption{Tatoeba results for each language. The languages are sorted according to their size in the pre-trained corpus from smallest to largest. Fine-tuning can significantly improve the accuracy of different language families in the cross-lingual retrieval task.}
\label{fig7}
\vskip -0.1in
\end{figure}

To further evaluate the performance of \textsc{Ernie-M} in retrieval task, we use hardest negative binary cross-entropy  loss \cite{wang2019camp,faghri2017vse++} to fine-tune \textsc{Ernie-M} with the same bilingual corpus in pre-training. Figure \ref{fig7} shows the details of accuracy on each language in Tatoeba. After fine-tuning, \textsc{Ernie-M} shows a significant improvement in all languages, with the average accuracy in all languages increasing from 87.9 to 93.3.

\subsection{Ablation Study}
To understand the effect of aligning semantic representations of multiple languages in the training process of \textsc{Ernie-M}, we conducted an ablation study as reported in Table \ref{table4}. $exp_{0}$ was directly fine-tuning XLM-R model on the XNLI and the CoNLL. We trained (1) only MMLM on the monolingual corpus, and the purpose of $exp_{1}$ was to measure how much performance gain could be achieved by continuing training based on the XLM-R model, (2) MMLM on the monolingual corpus, and TLM on the bilingual corpus, (3) MMLM on the monolingual corpus and CAMLM on the bilingual corpus, (4) MMLM and BTMLM on the monolingual corpus and CAMLM on the bilingual corpus and (5) full strategy of \textsc{Ernie-M}. We use the base model structure for our experiments, and to speed up the experiments, we use the XLM-R$_{\scriptsize \textsc{Base}}$ model to initialize the parameters of \textsc{Ernie-M}, all of which run 50,000 steps with the same hyperparameters with a batch size of 2048, and the score reported in the downstream task is the average score of five runs.

\renewcommand\arraystretch{0.9}
\begin{table}[!h]
\centering
\vskip 0.1in

\scalebox{0.7}{
\begin{tabular}{ccc|cc}
\toprule

\textbf{Index}& \textbf{Monolingual}& \textbf{Bilingual}& \textbf{XNLI}& \textbf{CoNLL}\\
\midrule
$exp_{0}$ & / & / & 75.7 & 79.2\\
$exp_{1}$ & MMLM & / & 75.8 & 79.2 \\
$exp_{2}$ & MMLM & TLM & 76.3 & 78.3\\
$exp_{3}$ & MMLM & CAMLM & 76.1 & 79.5\\
$exp_{4}$ & MMLM + BTMLM & CAMLM & 76.6 & 79.6\\
$exp_{5}$ & MMLM + BTMLM & CAMLM + TLM & \textbf{76.9} & \textbf{79.6}\\

\bottomrule
\end{tabular}}
\vskip -0.05in
\caption{Ablation study on each task in \textsc{Ernie-M}.}
\label{table4}
%\vskip -0in
\end{table}

\renewcommand\arraystretch{0.5}
\begin{table}[!h]
\centering
\vskip 0.1in

%\resizebox{\textwidth}{!}{
\scalebox{0.8}{
\begin{tabular}{lcc|c}
\toprule
%\midrule
\textbf{Model}& \textbf{MLQA}& \textbf{XNLI}& \textbf{Avg}\\
\midrule
mBERT & 23.3 & 16.9 & 20.1  \\
XLM-R & 17.6 & 10.4 & 14.0 \\
\textsc{InfoXLM} & 15.7 & 10.9 & 13.3 \\
\textsc{Ernie-M} & \textbf{15.0} & \textbf{8.8} & \textbf{11.9}\\
%\midrule
\bottomrule
\end{tabular}}
\vskip -0.05in
\caption{Cross-lingual transfer gap score, smaller gap indicates better transferability.}
\label{table5}
%\vskip -0.05in
\end{table}

Comparing $exp_{0}$ and $exp_{1}$, we can observer that there is no gain in the performance of the cross-lingual model by continuing pre-training XLM-R model. Comparing $exp_{2}$ $exp_{3}$ $exp_{4}$ with $exp_{1}$, we find that the learning of cross-lingual semantic alignment on parallel corpora is helpful for the performance of the model. Experiments that use the bilingual corpus for training show a significant improvement in XNLI. However, there are a surprised result that the using of TLM objective hurt the performance of NER task as $exp_{1}$ and $exp_{2}$ shows. Comparing $exp_{2}$ with $exp_{4}$, we find that our proposed BTMLM and CAMLM training objective are better for capturing the alignment of cross-lingual semantics. The training model with CAMLM and BTMLM objective results in a 0.3 improvement on XNLI and a 1.3 improvement on CoNLL compared to the training model with TLM. Comparing $exp_{3}$ to $exp_{4}$, we find that there is a 0.5 improvement on XNLI and 0.1 improvement on CoNLL after the model learns BTMLM. This demonstrates that our proposed BTMLM can learn cross-lingual semantic alignment and improve the performance of our model.

To further analyze the effect of our strategy, we trained the small-sized \textsc{Ernie-M} model from scratch. Table \ref{table5insert} shows the results of XNLI and CoNLL. Both XNLI and CoNLL results are the average of each languages. We observe that, \textsc{Ernie-M}$_{\scriptsize \textsc{Small}}$ can outperform XLM-R$_{\scriptsize \textsc{Small}}$ by 4.4 in XNLI and 6.6 in CoNLL. It suggests that our models can benefit from align cross-lingual semantic representation.

\begin{table}[!h]
\centering
\vskip 0.1in

%\resizebox{\textwidth}{!}{
\scalebox{0.8}{
\begin{tabular}{l|cc}
\toprule
%\midrule
\textbf{Model}& \textbf{CoNLL}& \textbf{XNLI}\\
\midrule
XLM-R & 63.2 & 55.7  \\
XLM-R + TLM & 65.6 & 67.3\\
XLM-R + CAMLM & 66.4 & 66.9 \\
XLM-R + CAMLM + BTMLM & 69.5 & 68.9 \\
\textsc{Ernie-M}$^{*}$ & 69.7 & 69.9\\
\textsc{Ernie-M} & \textbf{69.8} & \textbf{70.1}\\
%\midrule
\bottomrule
\end{tabular}}
\caption{XNLI and CoNLL accuracy under the cross-lingual transfer setting. All the models are small-sized trained from scratch. The small-sized model has the same hyperparameter as base model except that the number of layers is 6. \textsc{Ernie-M}$^{*}$ is the result in downstream tasks with the same computational overhead as XLM-R. All the models have the same training steps except \textsc{Ernie-M}$^{*}$.}
\label{table5insert}
\end{table}
%\vskip -0.3in

Table \ref{table5} shows the gap scores for English and other languages in the downstream task. This gap score is the difference between the English testset and the average performance on the testset in other languages. So, a smaller gap score represents a better transferability of the model. We can notice that the gap scores of \textsc{Ernie-M} are smaller compared to XLM-R and \textsc{InfoXLM} in both the XNLI and MLQA tasks, which indicates a better transferability of \textsc{Ernie-M}.

\begin{figure}[!htp]
\centering
\includegraphics[width=5.4cm]{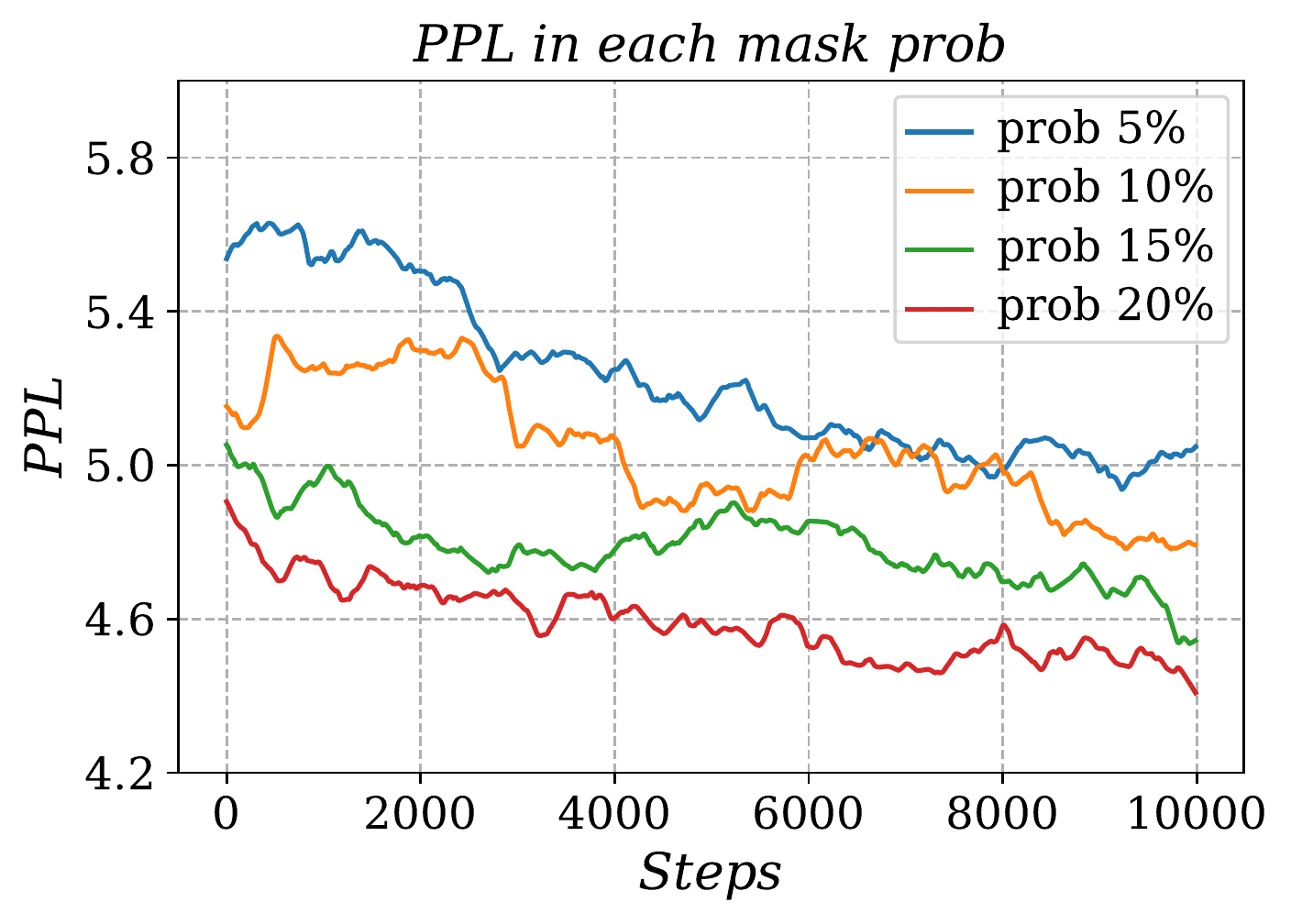}
\vskip -0.05in
\caption{PPL in BTMLM training with different mask prob, prob means the proportion of pseudo-tokens generated in BTMLM Stage 1.}
\label{fig6}
\end{figure}
\vskip -0.05in

To measure the computation cost of \textsc{Ernie-M}, we trained \textsc{Ernie-M} and XLM-R (MMLM + TLM) from scratch. 
The result shows that the training speed of \textsc{Ernie-M} is 1.075x compared with XLM-R, so the overall computational of \textsc{Ernie-M} is 1.075x compared with XLM-R.
With the same computational overhead, the performance of \textsc{Ernie-M} is 69.9 in XNLI and 69.7 in CoNLL, while XLM-R's performance is 67.3 in XNLI and 65.6 in CoNLL. The results demonstrate that \textsc{Ernie-M} performs better than XLM-R even with the same computational overhead.

In addition, we explored the effect of the number of generated pseudo-parallel tokens on the convergence of the model. In particular, we compare the impact on the convergence speed of the model when generating a 5\%, 10\%, 15\%, and 20\% proportion of pseudo-tokens. As shown in Figure \ref{fig6}, we can find that the perplexity (PPL) of the model decreases as the proportion of generated tokens increases, which indicates that the generated pseudo-parallel tokens are helpful for model convergence.

\section{Conclusion}
To overcome the constraint that the parallel corpus size places on the cross-lingual models performance, we propose a new cross-lingual model, \textsc{Ernie-M}, which is trained using both monolingual and parallel corpora. The contribution of \textsc{Ernie-M} is to propose two training objectives. The first objective is to enhance the multilingual representation on parallel corpora by applying CAMLM, and the second objective is to help the model to align cross-lingual semantic representations from a monolingual corpus by using BTMLM. Experiments show that \textsc{Ernie-M} achieves SoTA results in various downstream tasks on the XNLI, MLQA, CoNLL, PAWS-X, and Tatoeba datasets.

\bibliography{anthology,custom}
\bibliographystyle{acl_natbib}

%========================================

\clearpage
\appendix

\section{Appendix}

\subsection{Pre-training Data}
We follow \cite{wenzek2019ccnet} to reconstruct CC-100 data for \textsc{Ernie-M} training. The monolingual training corpus contains 96 languages, as shown in Table \ref{table6}. Note that several languages have the same ISO code, e.g., zh represents both Simplified Chinese and Traditional Chinese; ur represents both Urdu and Urdu Romanized. Table \ref{table7} shows the statistics of the parallel data in each language. 

\renewcommand\arraystretch{0.85}
\begin{table}[!h]
\centering
\vskip 0.1in

%\resizebox{\textwidth}{!}
\scalebox{0.75}{
\begin{tabular}{cc|cc|cc}
\toprule
\textbf{Code}& \textbf{Size (GB)}& \textbf{Code}& \textbf{Size (GB)}& \textbf{Code}& \textbf{Size (GB)}\\
\midrule
af & 0.1 & hi & 4.2 & or & 0.3 \\
am & 0.3 & hr & 1.0 & pa & 0.6 \\
ar & 12.5 & hu & 6.9 & pl & 20.2 \\
as & 0.1 & hy & 0.6 & ps & 0.3 \\
az & 0.6 & id & 11.7 & pt & 27.4 \\
be & 0.4 & is & 0.4 & ro & 7.5 \\
bg & 5.6 & it & 32.9 & ru & 215.6 \\
bn & 4.6 & ja & 78.1 & sa & 0.1 \\
br & 0.1 & jv & 0.1 & sd & 0.1 \\
bs & 0.1 & ka & 0.9 & si & 1.1 \\
ca & 2.1 & kk & 0.5 & sk & 9.5 \\
cs & 10.5 & km & 0.2 & sl & 4.3 \\
cy & 0.3 & kn & 0.2 & so & 0.1 \\
da & 4.8 & ko & 29.4 & sq & 2.0 \\
de & 71.0 & ku & 0.1 & sr & 5.5 \\
el & 10.5 & ky & 0.4 & su & 0.1 \\
en & 512.5 & la & 0.2 & sv & 42.1 \\
eo & 0.4 & lo & 0.2 & sw & 0.2 \\
es & 62.6 & lt & 1.7 & ta & 6.9 \\
et & 1.0 & lv & 0.9 & te & 2.0 \\
eu & 0.7 & mg & 0.1 & th & 29.1 \\
fa & 14.8 & mk & 0.5 & tl & 0.8 \\
fi & 4.3 & ml & 1.2 & tr & 43.3 \\
fr & 61.5 & mn & 0.3 & ug & 0.1 \\
fy & 0.1 & mr & 0.4 & uk & 11.1 \\
ga & 0.2 & ms & 0.5 & ur & 2.2 \\
gd & 0.1 & my & 0.4 & uz & 0.1 \\
gl & 1.0 & ne & 0.5 & vi & 52.0 \\
gu & 0.2 & nl & 17.8 & yi & 0.2 \\
he & 3.3 & no & 3.8 & zh & 96.0 \\

\bottomrule
\end{tabular}}
\caption{Statistics of CC-100 used for \textsc{Ernie-M} pre-training.}
\label{table6}
\end{table}
\vskip -0.3in

\begin{table}[!h]
\centering

\vskip 0.1in

\scalebox{0.75}{
\begin{tabular}{cc|cc}
\toprule

\textbf{ISO Code}& \textbf{Size (GB)}& \textbf{ISO Code}& \textbf{Size (GB)}\\
\midrule
ar & 9.8 & ru & 8.3 \\
bg & 2.2 & sw & 0.1 \\
de & 10.7 & th & 3.3 \\
el & 4.0 & tr & 1.1 \\
es & 8.8 & ur & 0.7 \\
fr & 13.7 & vi & 0.8 \\
hi & 0.3 & zh & 5.0 \\

\bottomrule
\end{tabular}}
\caption{Statistics of parallel data used for \textsc{Ernie-M} pre-training.}
\label{table7}
%\vskip -0.15in
\end{table}
\vskip -0.2in

\subsection{Hyperparameters for Pre-training}
Table \ref{table8} lists the hyperparameters for pre-training. We use the XLM-R model to initialize the parameters of base and large model, for the small model, we train it from scratch. The vocab of \textsc{Ernie-M} is the same as that of XLM-R.

\begin{table}[!h]
\centering

\vskip 0.1in

\scalebox{0.8}{
\begin{tabular}{llll}
\toprule
\textbf{Hyperparameters}& \textbf{\textsc{Small}}& \textbf{\textsc{Base}}& \textbf{\textsc{Large}} \\
\midrule
Layers & 6 & 12 & 24 \\
Hidden size & 768 & 768 & 1024 \\
FFN inner hidden size & 3,072 & 3,072 & 4,096 \\
FFN dropout & 0.1 & 0.1 & 0.1 \\
Attention heads & 12 & 12 & 16 \\
Attention dropout & 0.1 & 0.1 & 0.1 \\
Embedding size & 768 & 768 & 1024 \\
Training steps & 240K & 150K & 200K \\
Batch size & 1,024 & 2,048 & 2,048 \\
Learning rate & 3e-4 & 2e-4 & 1e-4 \\
Learning rate schedule & Linear & Linear & Linear \\
Adam $\varepsilon$ & 1e-6 & 1e-6 & 1e-6 \\
Adam $\beta_{1}$ & 0.98 & 0.98 & 0.98 \\
Adam $\beta_{2}$ & 0.999 & 0.999 & 0.999 \\
Weight decay & 0.01 & 0.01 & 0.01 \\
Warmup steps & 10,000 & 10,000 & 10,000 \\

\bottomrule
\end{tabular}}
\caption{Hyperparameters used for pre-training.}
\label{table8}
\vskip -0.2in
\end{table}

\subsection{Hyperparameters for Fine-tuning}
Tables \ref{table9} and \ref{table10} list the fine-tuning parameters on XNLI, MLQA, CoNLL and PAWS-X. For each task, we select the model with the best performance on the validation set, and the test set score is the average of five runs with different random seeds. Tables \ref{table13} list the fine-tuning parameters on Tatoeba.

\renewcommand\arraystretch{0.9}
\begin{table*}[!ht]
\centering
%\captionsetup{width=.75\textwidth}

\vskip 0.1in

\scalebox{0.85}{
\begin{tabular}{lccccc}
\toprule

\textbf{Hyperparameters}& \textbf{XNLI}&  \textbf{XNLI}$^{*}$ & \textbf{MLQA} & \textbf{CoNLL} & \textbf{CoNLL}$^{*}$ \\
\midrule
Batch size & 32 & 128 & 32 & 8 & 8\\
Learning rate & 5e-5 & 5e-5 & 3e-4 & 4e-4 & 3e-4\\
Layerwise LR decay & 0.8 & 0.8 & 0.8 & 0.8 & 0.8\\
LR schedule & Linear & Linear & Linear & Linear & Linear \\
Warmup faction & 10\% & 10\% & 10\% & 10\% & 10\% \\
Weight decay & 0 & 0 & 0 & 0.01 & 0.01 \\
Epoch & 5 & 2 & 2 & 10 & 10\\

\bottomrule
\end{tabular}}
\caption{Hyperparameters used for $\textsc{Ernie-M}_{\scriptsize \textsc{Small}}$ and $\textsc{Ernie-M}_{\scriptsize \textsc{Base}}$ fine-tuning; parameters with ``*'' are in the translate-train-all setting, and those without ``*'' are in the cross-lingual setting. }
\label{table9}
\end{table*}

\renewcommand\arraystretch{0.9}
\begin{table*}[!ht]
\centering
%\captionsetup{width=.75\textwidth}

\vskip 0.1in

\scalebox{0.85}{
\begin{tabular}{lccccccc}
\toprule

\textbf{Hyperparameters}& \textbf{XNLI}&  \textbf{XNLI}$^{*}$ & \textbf{MLQA} & \textbf{CoNLL} & \textbf{CoNLL}$^{*}$  & \textbf{PAWS-X} & \textbf{PAWS-X}$^{*}$\\
\midrule
Batch size & 32 & 128 & 32 & 8 & 8 & 64 & 64\\
Learning rate & 5e-5 & 5e-5 & 8e-5 &  4e-4 & 3e-4 & 5e-5 & 7e-5\\
Layerwise LR decay & 0.8 & 0.8 & 0.9 & 0.8 & 0.8 & 0.9 & 0.9\\
LR schedule & Linear & Linear & Linear & Linear & Linear & Linear & Linear\\
Warmup faction & 10\% & 10\% & 10\% & 10\% & 10\% & 10\% & 10\%\\
Weight decay & 0 & 0 & 0 & 0.01 & 0.01 & 0.01 & 0.01\\
Epoch & 5 & 1 & 2 & 10 & 10 & 10 & 2\\

\bottomrule
\end{tabular}}
\caption{Hyperparameters used for $\textsc{Ernie-M}_{\scriptsize \textsc{Large}}$ fine-tuning;  parameters with ``*'' are in the translate-train-all setting, and those without ``*'' are in the cross-lingual setting.}
\label{table10}
\end{table*}

%%%%%%%%%%%%%%%%%%
\begin{table*}[!h]
\centering

\vskip 0.1in

\scalebox{0.85}{
\begin{tabular}{lll}
\toprule
\textbf{Hyperparameters}& \textbf{\textsc{Large}} \\
\midrule
Training steps & 200K  \\
Batch size & 32  \\
Learning rate & 5e-5 \\
Learning rate schedule & Linear  \\
Weight decay & 0.0 \\
Warmup faction & 10\% \\
\bottomrule
\end{tabular}}
\caption{Hyperparameters used for $\textsc{Ernie-M}_{\scriptsize \textsc{Large}}$ fine-tuneing in Tatoeba.}
\label{table13}
\end{table*}
%%%%%%%%%%%%%%%%%%

\subsection{Results for 15 languages model}

To better evaluate the performance of \textsc{Ernie-M}, we train the \textsc{Ernie-M}-15 model for 15 languages. The languages of training corpora is the same as that of \textsc{Hictl} \cite{wei2020learning}. We evaluate \textsc{Ernie-M}-15 on the XNLI dataset. Table \ref{table11} shows the results of 15 languages models. The \textsc{Ernie-M}-15 model outperforms the current best 15-language cross-lingual model on the XNLI task, achieving a score of 77.5 in the cross-lingual transfer setting, outperforming \textsc{Hictl} 0.2 and a score of 80.7 in the translate-train-all setting, outperforming \textsc{Hictl} 0.7. 

\renewcommand\arraystretch{0.9}
\begin{table*}[!htb]
\centering

\vskip 0.1in

\resizebox{\textwidth}{!}{

\begin{tabular}{l|ccccccccccccccc|c}
\toprule
\textbf{Model}& \textbf{en}& \textbf{fr}& \textbf{es}& \textbf{de}& \textbf{el}& \textbf{bg}& \textbf{ru}& \textbf{tr}& \textbf{ar}& \textbf{vi}& \textbf{th}& \textbf{zh}& \textbf{hi}& \textbf{sw}& \textbf{ur}& \textbf{Avg}  \\
\midrule
\multicolumn{17}{l}{\textit{Fine-tune cross-lingual model on English training set (Cross-lingual Transfer)}} \\
\midrule
XLM \cite{lample2019cross}& 85.0 & 78.7 & 78.9 & 77.8 & 76.6 & 77.4 & 75.3 & 72.5 & 73.1 & 76.1 & 73.2 & 76.5 & 69.6 & 68.4 & 67.3 & 75.1 \\
\textsc{Hictl} \cite{wei2020learning}& \textbf{86.3} & 80.5 & 81.3 & 79.5 & 78.9 & 80.6 & \textbf{79.0} & 75.4 & 74.8 & 77.4 & 75.7 & \textbf{77.6} & \textbf{73.1} & \textbf{69.9} & \textbf{69.7} & 77.3 \\
\textsc{Ernie-M}-15 & 85.9 & \textbf{80.5} & \textbf{81.3} & \textbf{79.8} & \textbf{79.3} & \textbf{80.7} & 78.7 & \textbf{76.8} & \textbf{76.8} & \textbf{78.0} & \textbf{76.1} & 77.4 & 72.9 & 68.9 & 68.9 & \textbf{77.5} \\

\midrule
\multicolumn{17}{l}{\textit{Fine-tune cross-lingual model on all training sets (Translate-Train-All)}} \\
\midrule
XLM \cite{lample2019cross}& 85.0 & 80.8 & 81.3 & 80.3 & 79.1 & 80.9 & 78.3 & 75.6 & 77.6 & 78.5 & 76.0 & 79.5 & 72.9 & 72.8 & 68.5 & 77.8 \\
% \midrule
\textsc{Hictl} \cite{wei2020learning}& \textbf{86.5} & 82.3 & 83.2 & 80.8 & 81.6 & 82.2 & \textbf{81.3} & 80.5 & 78.1 & 80.4 & 78.6 & \textbf{80.7} & 76.7 & 73.8 & \textbf{73.9} & 80.0 \\
\textsc{Ernie-M}-15 & 86.4 & \textbf{82.4} & \textbf{83.5} & \textbf{82.7} & \textbf{83.1} & \textbf{83.2} & 81.0 & \textbf{80.6} & \textbf{80.5} & \textbf{80.9} & \textbf{79.2} & 80.6 & 77.7 & \textbf{75.8} & 72.8 & \textbf{80.7} \\
\bottomrule
\end{tabular}}
\caption{Evaluation results on XNLI cross-lingual natural language inference for 15 languages model.}
\label{table11}
\end{table*}

\renewcommand\arraystretch{0.9}
\begin{table*}[!htb]
\centering

\vskip 0.1in

\resizebox{\textwidth}{!}{

\begin{tabular}{l|cccccccccccccccccc}
\toprule
\textbf{Model}& \textbf{af}& \textbf{ar}& \textbf{bg}& \textbf{bn}& \textbf{de}& \textbf{el}& \textbf{es}& \textbf{et}& \textbf{eu}& \textbf{fa}& \textbf{fi}& \textbf{fr}& \textbf{he}& \textbf{hi}& \textbf{hu}& \textbf{id}& \textbf{it}& \textbf{ja} \\
\midrule
VECO$_{\scriptsize \textsc{Large}}$ \cite{luo2020veco}& 80.9& 85.1& 91.3& 78.1& 98.5& 89.5& \textbf{97.4}& \textbf{94.8}& \textbf{79.8}& \textbf{93.1}& \textbf{95.4}& 93.7& 85.8& 94.2& \textbf{93.8}& \textbf{93.0}& \textbf{92.2}& \textbf{92.8} \\
\textsc{Ernie-M}$_{\scriptsize \textsc{Large}}$ & \textbf{88.6}& \textbf{88.9}& \textbf{92.2}& \textbf{84.8}& \textbf{98.8}& \textbf{92.4}& 96.3& 82.4& 78.8& 92.5& 92.2& \textbf{94.0}& \textbf{86.2}& \textbf{95.4}& 88.7& 91.5& 90.3& 86.7 \\
\hline
\textsc{Ernie-M}$^\dag_{\scriptsize \textsc{Large}}$ & 92.6 & 94.3 & 96.6 & 89.2 & 99.7 & 96.8 & 98.8 & 92.5 & 87.4 & 96.0 & 97.1 & 96.5 & 90.1 & 97.9 & 95.5 & 95.7 & 95.2 & 96.9 \\
\midrule
\textbf{Model}& \textbf{jv}& \textbf{ka}& \textbf{kk}& \textbf{ko}& \textbf{ml}& \textbf{mr}& \textbf{nl}& \textbf{pt}& \textbf{ru}& \textbf{sw}& \textbf{ta}& \textbf{te}& \textbf{th}& \textbf{tl}& \textbf{tr}& \textbf{ur}& \textbf{vi}& \textbf{zh}  \\
\midrule
% \midrule
VECO$_{\scriptsize \textsc{Large}}$ \cite{luo2020veco}& 35.1& 83.0& 74.1& \textbf{88.7}& 94.8& 82.5& 95.9& \textbf{94.6}& 92.2& \textbf{69.7}& 82.4& 91.0& 94.7& 73.0& 95.2& 63.8& \textbf{95.1}& \textbf{93.9} \\
\textsc{Ernie-M}$_{\scriptsize \textsc{Large}}$ & \textbf{48.0}& \textbf{84.2}& \textbf{78.2}& 85.3& \textbf{95.2}& \textbf{87.6}& \textbf{96.1}& 92.6& \textbf{93.1}& 59.4& \textbf{86.9}& \textbf{94.0}& \textbf{95.6}& \textbf{75.4}& \textbf{96.3}& \textbf{90.8}& 94.5& 91.7  \\
\hline
\textsc{Ernie-M}$^\dag_{\scriptsize \textsc{Large}}$ & 65.2 & 94.9 & 88.0 & 94.1 & 98.5 & 90.8 & 98.1 & 94.5 & 95.7 & 68.4 & 91.8 & 97.9 & 98.4 & 86.0 & 98.3 & 94.9 & 98.1 & 96.7 \\
\bottomrule
\end{tabular}}
\caption{Tatoeba results for each language. ``$\dag$'' indicates the results after fine-tuning}
\label{table12}
\end{table*}

\subsection{Results for Cross-lingual Retrieval}

Table \ref{table12} shows the details of accuracy on each language in the cross-lingual retrieval task. For a fair comparison with VECO, we use the averaged representation in the middle layer of best XNLI model for cross-lingual retrieval task. \textsc{Ernie-M} outperforms VECO in most languages and achieves state-of-the-art results. We also proposed a new method for cross-lingual retrieval. We use hardest negative binary cross-entropy loss \cite{wang2019camp,faghri2017vse++} to fine-tune \textsc{Ernie-M} with the same bilingual corpora in pre-training. Table \ref{table12} report the results after fine-tuning, the average accuracy of Tatoeba improve from 87.9 to 93.3.

\end{document}